%% file: acl_latex.tex
\pdfoutput=1

\documentclass[11pt]{article}

\usepackage[]{acl}
\usepackage{balance}

\usepackage{times}
\usepackage{latexsym}
\usepackage{graphicx}
\usepackage[T1]{fontenc}

\usepackage[utf8]{inputenc}

\usepackage{microtype}
\usepackage{subfigure}
\usepackage{inconsolata}
\usepackage{multirow}
\usepackage{amsmath}
\usepackage{booktabs}
\usepackage{algorithm}
\usepackage{algpseudocode}

\definecolor{midnightgreen}{rgb}{0.0, 0.29, 0.33}

\title{Cleaner Pretraining Corpus Curation with Neural Web Scraping}

\author{Zhipeng Xu$^{1}$, Zhenghao Liu$^{1}$\thanks{ \ \ indicates corresponding author.}, Yukun Yan$^{2}$, Zhiyuan Liu$^{2}$, Ge Yu$^{1}$ and Chenyan Xiong$^{3}$\\ 
$^1$Department of Computer Science and Technology, Northeastern University, China \\
$^2$Department of Computer Science and Technology, Institute for AI, Tsinghua University, China \\
Beijing National Research Center for Information Science and Technology, China \\
$^3$Language Technologies Institute, Carnegie Mellon University, United States\\
}

\begin{document}
\maketitle

\input{section/0_abstract}

\input{section/1_introduction}

\input{section/2_related}

\input{section/3_methodology}

\input{section/4_experiment}

\input{section/5_result}

\input{section/6_conclusion}

\input{section/limitation}


\bibliography{custom}
\clearpage
\newpage
\appendix
\input{section/7_appednix}

\end{document}

%% file: section/0_abstract.tex
\begin{abstract}
The web contains large-scale, diverse, and abundant information to satisfy the information-seeking needs of humans. Through meticulous data collection, preprocessing, and curation, webpages can be used as a fundamental data resource for language model pretraining. However, when confronted with the progressively revolutionized and intricate nature of webpages, rule-based/feature-based web scrapers are becoming increasingly inadequate. This paper presents a simple, fast, and effective \textbf{Neu}ral web \textbf{Scraper} (\texttt{NeuScraper}) to help extract primary and clean text contents from webpages. Experimental results show that \texttt{NeuScraper} surpasses the baseline scrapers by achieving more than a 20\% improvement, demonstrating its potential in extracting higher-quality data to facilitate the language model pretraining.  All of the code is available at \url{https://github.com/OpenMatch/NeuScraper}.
\end{abstract}

%% file: section/1_introduction.tex
\section{Introduction}
Large Language Models (LLMs) have shown impressive performance in various NLP tasks as the size of models scaling up~\cite{chowdhery2023palm, touvron2023llama, achiam2023gpt, zhao2023survey}. However, recent findings in scaling laws indicate that both model size and training data should be scaled proportionally~\cite{hoffmann2022training}, posing a significant challenge in acquiring sufficiently large pretraining datasets or even raising concerns about data scarcity~\cite{penedo2023refinedweb,villalobos2022will}.


To curate more data for pretraining, researchers pay more attention to collecting more valuable data from the Web. The web-crawled datasets, such as CommonCrawl, have been widely used for pretraining, facilitating the development of language models~\cite{wenzek2019ccnet,radford2019language,raffel2020exploring,penedo2023refinedweb}. Nevertheless, prior research has demonstrated that, even after aggressive cleaning, the quality of pre-extracted text provided by CommonCrawl still fails to reach the expected~\cite{raffel2020exploring,gao2020pile,penedo2023refinedweb}. The reason lies in that advertisements, banners, hyperlinks, and other harmful content are usually mixed within the primary content of the page, thereby only extracting these primary contents brings lots of noise to pretraining~\cite{gibson2005volume,vogels2018web2text}.

Web scrapers provide opportunities to extract valuable content from the raw HTML pages~\cite{barbaresi2021trafilatura}. However, rule-based and heuristic scrapers have notable limitations. On the one hand, web pages are becoming increasingly sophisticated, requiring more intricate underlying code to deal with the page layout~\cite{butkiewicz2011understanding}. In this case, maintaining the scraper rules is time-consuming and requires much human effort. On the other hand, HTML functions as a markup language, enabling web designers to customize web pages according to individual preferences. Consequently, web pages frequently lack complete standardization, which may mislead the rule-based web scrapers~\cite{hantke2022html}.

In this paper, we present a simple, fast, and effective Neural Web Scraper (\texttt{NeuScraper}) designed to extract primary content from webpages. \texttt{NeuScraper} employs a shallow neural architecture and integrates layout information for efficient scraping. Our experiments demonstrate that \texttt{NeuScraper} surpasses baseline scrapers, achieving a 20\% improvement in performance and generating a higher-quality corpus for language model pretraining. Notably, \texttt{NeuScraper} shows the potential of high processing speeds when utilized on GPU. The easy-to-use and open-source tool, \texttt{NeuScraper}, can facilitate the creation of large-scale corpora for pretraining.

%% file: section/2_related.tex
\section{Related Work}
Leveraging web scrapers for extraction provides a promising way to extract high-quality content from webpages. Such a web scraping task is usually defined as text extraction, boilerplate removal, template removal, or generic web extraction in different webpage processing pipelines~\cite{finn2001fact,rahman2001content,vieira2006fast}, which is distinguished from the web information extraction task that extracts the entities from webpages~\cite{li2021markuplm,wang2022webformer}. The web scrapers can be divided into rule-based and feature-based methods.

\input{figure/newparser}
Rule-based web scrapers start from web wrappers, which often need manual designs or a wrapper induction system for producing~\cite{muslea1999hierarchical,crescenzi2001roadrunner}. The web wrappers usually need to be tailored for each webpage, which is not feasible to process large-scale webpages~\cite{guo2010econ}. A more conventional approach is to create a Document Object Model (DOM) tree, which assists in building a rule-based scraper~\cite{gupta2003dom,guo2010econ} or help the comparison of webpages~\cite{yi2003eliminating}. Additionally, the work also incorporates tag cumulative distributions~\cite{finn2001fact}, text density~\cite{sun2011dom}, and tag ratios~\cite{weninger2010cetr} to benefit the content extraction from webpages.

Except for these rule-based methods, some scrapers use feature-based approaches to better extract the primary contents from webpages. Specifically, they divide the webpage into several blocks using rules built based on the HTML tags or DOM tree. Then they extract dozens to hundreds of hand-crafted features from these blocks, such as markup, text/document features~\cite{spousta2008victor}, linguistic, structural \& visual features~\cite{bauer2007fiasco} and DOM tree-based features~\cite{vogels2018web2text}. These features can be fed into SVM~\cite{bauer2007fiasco,kohlschutter2010boilerplate}, conditional random fields~\cite{spousta2008victor}, logistic regressions~\cite{peters2013content} or convolutional neural network~\cite{vogels2018web2text} to classify whether the texts in the block are the primary content of the webpages.

%% file: figure/newparser.tex
\begin{figure}[t]
  \centering
  \includegraphics[width=0.48\textwidth]{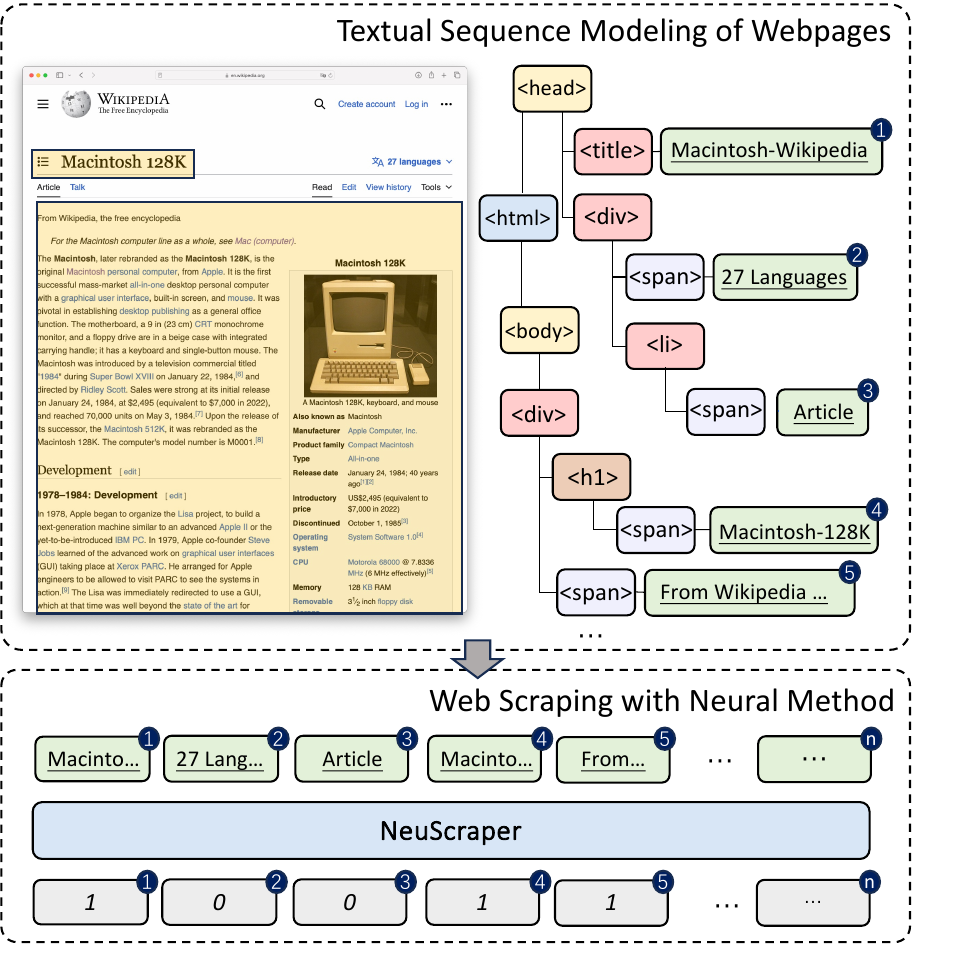}
  \caption{The Pipeline of Primary Content Extraction Using \texttt{NeuScraper} (Neural Web Scraper).}
  \label{fig:model}
\end{figure}

%% file: section/3_methodology.tex
\section{Neural Web Scraper}

This section introduces our Neural Web Scraper (\texttt{NeuScraper}) to extract primary contents from webpages. We first introduce the sequence modeling method of webpages (Sec.~\ref{model:task}) and then describe our neural-based web scraper (Sec.~\ref{model:model}).

\subsection{Textual Sequence Modeling of Webpages}\label{model:task}
As shown in Figure~\ref{fig:model}, the primary content extraction task aims to extract the content from the highlighted areas, which consists of clean text and represents the main information of the webpage. To facilitate the web scraping with \texttt{NeuScraper}, we convert the HTML code into textual sequences.

Previous work~\cite{bauer2007fiasco} has demonstrated the effectiveness of both structural and visual features in helping to identify primary contents. Thus, to preserve webpage layout information, we rely on the DOM tree structure to transform webpages into textual sequences. Specifically, we employ the BeautifulSoup4\footnote{\url{ https://pypi.org/project/beautifulsoup4/}} toolkit to build the DOM tree for each webpage, conduct the depth-first traversal on the tree and regard the visited order as additional location information to represent the nodes. During this process, only the nodes that contain plant texts, table nodes (tagged with \textit{<table>}), and list nodes (tagged with \textit{<ol>}, \textit{<ul>} or \textit{<dl>}) are reserved to produce the final textual sequences $X= \{x_1, x_2, ..., x_n\}$, where $n$ denotes the number of the reserved DOM nodes. After processing, the web scraping task primarily involves determining whether the node $x_i$ contains the primary content of the webpage for evaluation.

\subsection{Web Scraping with the Neural Method}\label{model:model}

In this subsection, we introduce our neural modeling method to build the web scraper.
To process the textual sequences $X = \{x_1, x_2, ..., x_n\}$, we build a hierarchical architecture for node-level prediction.

Specifically, to guarantee the efficiency of \texttt{NeuScraper}, we use the first layer of the XLMRoberta~\cite{conneau2019unsupervised} model to encode the text representation $x_i$ of the $i$-th DOM node as the 768-dimensional node representation $h_i$:
\begin{equation}
\small
    h_i = \text{XLMRoberta-Layer}^1 (x_i),
\end{equation}
where $h_i$ is the representation of the ``[CLS]'' token. Then we feed these node representations $H=\{h_1, h_2,...,h_n\}$ into a 3-layer transformer model~\cite{vaswani2017attention} with 8 attention heads to get the encoded node representations:
\begin{equation}
\small
    \hat{h}_i = \text{Transformer} (\text{Linear}(h_i)),
\end{equation}
where the linear layer projects $h_i$ to 256-dimensional embeddings for efficient modeling. Following previous work~\cite{overwijk2022clueweb22}, the DOM nodes can be categorized into six kinds of labels $y^k$, including primary content, heading, title, paragraph, table, and list. Then we calculate the label prediction probability $P(y_i^k=1|x_i)$ of the $k$-th category label $y_i^k$ of the $i$-th node:
\begin{equation}
\small
    P(y_i^k=1|x_i) = \text{Sigmoid} (\text{MLP} (\hat{h}_i))
\end{equation}
Finally, \texttt{NeuScraper} is trained using the loss $L$:
\begin{equation}
\small
    L = \sum_{k=1}^6 \sum_{i=1}^n \text{CrossEntropy} (P(y_i^k|x_i), \mathcal{Y}_i^k),
\end{equation}
where $\mathcal{Y}_i^k$ is the ground truth label. $\mathcal{Y}_i^k$ is a binary label and $\mathcal{Y}_i^k=1$ indicates that the $i$-th DOM node belongs to the $k$-th label category. During inference, we only consider the primary content label to extract the texts from webpages.

%% file: section/4_experiment.tex
\section{Experimental Methodology}

In this section, we describe the datasets, baselines, evaluation metrics
and implementation details.

\textbf{Dataset.}
We use ClueWeb22~\cite{overwijk2022clueweb22} dataset in experiments. The content extraction labels of ClueWeb22 were generated from the production system of a commercial search engine. The labels are not available for general web scraping tools, because they are annotated with more expensive signals of page rendering and visualization. More details are shown in Appendix~\ref{app:overall}.

\textbf{Baseline.}
The scraping baselines consist of nine open-sourced web scrapers, including basic HTML manipulators (\texttt{html2text} and \texttt{inscriptis}~\cite{weichselbraun2021inscriptis}), generic webpage parsers (\texttt{beautifulsoup4}, \texttt{lxml} and \texttt{htmlparser}), rule-based scrapers (\texttt{jusText} and \texttt{readability}) and machine learning-based scraper (\texttt{boilerpipe}~\cite{kohlschutter2010boilerplate}). \texttt{trafilatura}~\cite{barbaresi2021trafilatura} is our main baseline, which combines different rules and heuristic methods. 

\textbf{Evaluation Metrics.}
The accuracy, precision, recall, and F1 score, are used to evaluate the effectiveness in extracting primary contents. Furthermore, we use different scrapers to produce the web corpus and pretrain language models. The quality of scraping can be demonstrated by the results of standard downstream tasks.

\textbf{Implementation Details.}
\texttt{NeuScraper} is trained for 30 epochs using the AdamW optimizer with a batch size of 1024. Learning rate adjustments followed the cosine decay schedule, with a warm-up phase spanning the initial 5\% of iterations and the peak rate fixed at 6e-4. To accommodate memory and computational speed limitations, the maximum length of node sequences was chunked to 384.

%% file: section/5_result.tex
\input{table/scraping}
\input{table/pretraining}
\section{Evaluation Result}
In this section, we first show the effectiveness of different scrapers in extracting primary contents from the raw webpages. Subsequently, we evaluate the quality of the extracted data and utilize it to pretrain language models of varying scales.

\subsection{Overall Performance}
The effectiveness of baseline scrapers and our \texttt{NeuScraper} in extracting primary contents from the raw webpages is shown in Table~\ref{tab:overall}.
Among all baseline scrapers, the \texttt{trafilatura} exhibits the highest performance, showcasing its effectiveness in content extraction through its cascade of rule-based filters and content heuristic methods. Our \texttt{NeuScraper} surpasses all traditional web scrapers and achieves over a 20\% improvement. It illustrates the effectiveness of our \texttt{NeuScraper} in learning the schemes of the primary contents, generalizing its advances to handle various layouts of pages and extracting high-quality texts from them. Notably, with the GPU support and distributed computation, \texttt{NeuScraper} achieves competitive scraping latency.

\subsection{Effectiveness of the Cleaned Web Data in Language Model Pretraining}\label{sec:pretraining}
This part evaluates the effectiveness of language models pretrained on the web data.

As shown in Table~\ref{tab:pretrain}, we utilize different scrapers to handle the webpages sourced from ClueWeb22 and CommonCrawl, and leverage the extracted data to pretrain Pythia models~\cite{biderman2023pythia}. The evaluation results demonstrate that employing the \texttt{NeuScraper} for webpage processing enhances the performance of language models in downstream tasks. It is noteworthy that the \texttt{NeuScraper} represents a data-driven scraping approach, circumventing the need for building sophisticated rules and conducting intricate feature engineering to deal with the continuously evolving HTML layouts.

\input{figure/ppl}
\subsection{Evaluation on the Quality of Extracted Data Using NeuScraper}
In this subsection, we aim to estimate the quality of extracted data using \texttt{NeuScraper}. The evaluation results are shown in Figure~\ref{fig:perplexity}.

It is apparent that if two corpora are of comparable quality, their n-gram distributions should exhibit similarity. Thus, we use the language models pretrained on web data (the same as Sec.~\ref{sec:pretraining}) to ask these language models to reproduce the target corpora, such as Wikitext~\cite{merity2016pointer} and Lambada~\cite{radford2019language}. The perplexity is used to evaluate the effectiveness of the language models pretrained on web data in replicating the target corpora. The lower perplexity indicates the language model is more proficient to the target corpora, showing the pretrained data and target data have more overlaps and are more similar.

The evaluation results reveal that the utilization of extracted content from some simple scrapers, such as \texttt{htmlparser}, significantly impacts the effectiveness of language models, which causes an increase of more than 20 points in perplexity due to the noise derived from webpages. Compared with the \texttt{trafilatura}, \texttt{NeuScraper} decreases the perplexity by over ten points, showing its capability to yield higher-quality data for pretraining through learning to extract primary content.

\subsection{Model Quantization for NeuScraper}
In this subsection, we quantize the model of \texttt{NeuScraper} via \texttt{onnxruntime}\footnote{\url{https://onnxruntime.ai}} to evaluate its efficiency in resource-constrained scenarios. 

As shown in Table~\ref{tab:quantize}, we utilize \texttt{qint8} and \texttt{quint8} to quantize our \texttt{NeuScraper}. The \texttt{qint8} quantizes model parameters or layer outputs to signed 8-bit integers, while \texttt{quint8} quantizes them to unsigned 8-bit integers, reducing model size and improving computational efficiency. Benefiting from quantization, \texttt{NeuScraper} accelerates by 25\% with no loss of performance compared to the original model. While processing is still 4-5x slower compared to GPUs, it also provides a potential way to scrap in low-resource scenarios via \texttt{NeuScraper}. 

\input{table/quantization}

%% file: table/scraping.tex
\begin{table}[t]
\centering
\resizebox{\linewidth}{!}{
\begin{tabular}{l|cccc|c}
    \hline
    \multirow{2}{*}{\textbf{Method}}&\multicolumn{4}{c|}{\textbf{Evaluation Metrics}}&\multicolumn{1}{c}{\textbf{Latency}} \\ 
    &\textbf{Acc.}&\textbf{Prec.}&\textbf{Rec.}&\textbf{F1}&\multicolumn{1}{c}{\textbf{(ms)}} \\ \hline
        htmlparser&40.73&40.65&98.95&57.63&19.01\\
        bs4&41.29&40.96&\textbf{99.94}&58.10&12.65\\
        html2text&40.44&39.35&85.40&53.88&15.85\\
        boilerpipe&66.48&66.79&35.27&46.16&11.05\\
        jusText&62.58&72.62&13.08&22.17&10.91\\
        lxml&64.62&61.48&35.22&44.78&10.96\\
        inscriptis&45.35&42.48&96.43&58.98&14.99\\
        readability&68.47&72.84&36.04&48.22&12.36\\
        trafilatura&70.70&66.57&56.42&61.08&11.95\\ \hline
        NeuScraper&\textbf{86.35}&\textbf{80.77}&87.29&\textbf{83.90}&11.39\\
    \hline
    \end{tabular}}
    \caption{Overall Performance. We use ClueWeb22 to evaluate the content extraction effectiveness of different web scrapers. More details are shown in Appendix~\ref{app:overall}.}
    \label{tab:overall}
\end{table}

%% file: table/pretraining.tex
\begin{table*}[]
\centering
\resizebox{\linewidth}{!}{
\begin{tabular}{l|l|cccccccc|c}
    \hline
\textbf{Size}&\textbf{Method}&\textbf{BLIMP}&\textbf{ARC-e}&\textbf{ARC-c}&\textbf{SWAG}&\textbf{WinoG}&\textbf{SciQ}&\textbf{Lambada}&\textbf{LogiQA} &\textbf{AVG} \\ \hline
        \multicolumn{5}{l}{\textbf{\textit{ClueWeb22}}} \\ \hline
        \multirow{3}{*}{160M}
        &htmlparser&70.87&41.16&17.23&32.24&49.88&66.10&16.96&22.58&39.63 \\
        &trafilatura&73.46&42.46&18.25&34.08&48.61&\textbf{69.20}&\textbf{18.10}&\textbf{22.11}&40.78\\
        &NeuScraper&\textbf{74.01}&\textbf{42.84}&\textbf{18.43}&\textbf{34.14}&\textbf{51.46}&69.00&17.58&21.50&\textbf{41.12}\\ \hline
        \multirow{3}{*}{410M}
        &htmlparser&74.24&42.63&18.77&34.45&49.80&70.80&22.35&22.42&41.93 \\
        &trafilatura&\textbf{77.84}&45.28&\textbf{20.56}&\textbf{37.29}&\textbf{52.32}&72.90&23.77&21.96&43.99\\
        &NeuScraper&76.71&\textbf{47.34}&20.47&37.00&50.74&\textbf{74.40}&\textbf{26.76}&\textbf{24.42}&\textbf{44.73}\\ \hline
    \multicolumn{5}{l}{\textbf{\textit{CommonCrawl}}} \\ \hline
        \multirow{3}{*}{160M}
        &htmlparser&58.38&29.71&\textbf{18.77}&28.85&\textbf{50.27}&38.60&5.16&19.66&31.17 \\
        &trafilatura&\textbf{69.72}&34.72&18.51&32.04&49.56&56.90&11.70&\textbf{23.96}&37.13 \\
        &NeuScraper&69.27&\textbf{36.15}&18.43&\textbf{32.61}&\textbf{51.77}&\textbf{60.50}&\textbf{15.48}&20.73&\textbf{38.12} \\ \hline
        \multirow{3}{*}{410M}
        &htmlparser&61.30&28.28&17.23&29.36&50.35&41.00&6.50&20.73&31.84 \\
        &trafilatura&72.66&36.74&\textbf{20.13}&33.91&\textbf{51.30}&55.40&16.08&21.35&38.44 \\
        &NeuScraper&\textbf{74.42}&\textbf{39.30}&18.60&\textbf{34.77}&50.03&\textbf{61.40}&\textbf{20.66}&\textbf{21.81}&\textbf{40.12}\\ 

    \hline
    \end{tabular}}
    \caption{Effectiveness of Pythia Pretraining Using the Extracted Data from Different Scrapers. We pretrained Pythia models of different sizes on ClueWeb22 and CommonCrawl respectively. More details are shown in Appendix~\ref{app:pretrain}. }
    \label{tab:pretrain}
\end{table*}

%% file: figure/ppl.tex
\begin{figure}[t]
    \centering
    \subfigure[ClueWeb22.]{ \label{fig:ppl_cw} 
    \includegraphics[width=0.48\linewidth]{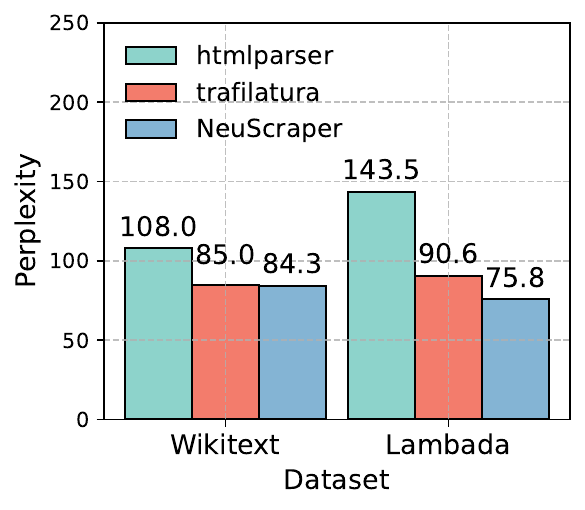}}
    \subfigure[CommonCrawl.]{ \label{fig:ppl_cc} 
    \includegraphics[width=0.48\linewidth]{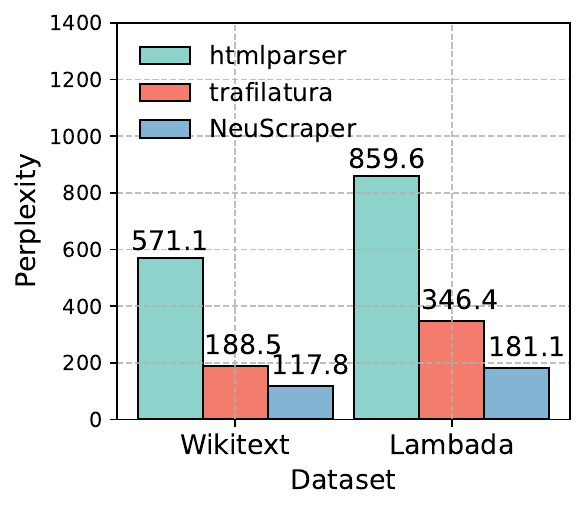}}
    \caption{The Effectiveness of Language Models Trained on Web Data to Reproduce the Target Corpora. Lower perplexity indicates more proficiency in language models for reproducing.}
    \label{fig:perplexity}
\end{figure}

%% file: table/quantization.tex
\begin{table}[t]
\centering
\resizebox{\linewidth}{!}{
\begin{tabular}{l|cccc|c}
    \hline
    \multirow{2}{*}{\textbf{Method}}&\multicolumn{4}{c|}{\textbf{Evaluation Metrics}}&\multicolumn{1}{c}{\textbf{Latency}} \\ 
    &\textbf{Acc.}&\textbf{Prec.}&\textbf{Rec.}&\textbf{F1}&\multicolumn{1}{c}{\textbf{(ms)}} \\ \hline
CPU&86.35&80.77&87.29&83.90&55.25 \\
 + \texttt{qint8}&86.37&80.70&87.48&83.95&42.22 \\ 
 + \texttt{quint8}&86.39&80.68&87.56&83.98&41.48 \\\hline
GPU&86.35&80.77&87.29&83.90&11.39 \\\hline
    \end{tabular}}
    \caption{Quantization Performance of \texttt{NeuScraper} on ClueWeb22. We further quantized \texttt{NeuScraper} to accelerate its inference on the CPU.}
    \label{tab:quantize}
\end{table}

%% file: section/6_conclusion.tex
\section{Conlusion}
This paper proposes \texttt{NeuScraper}, which employs a shallow neural architecture to clean the webpages. The experimental results show the effectiveness of \texttt{NeuScraper}. The open-sourced and easy-used web scraper may facilitate the research on language model pretraining.

%% file: section/limitation.tex
\section*{Limitation}
To guarantee efficiency, \texttt{NeuScraper} needs the powerful parallelism of GPUs to achieve high-speed web scraping. In addition, for large-scale pretraining corpus processing, a high throughput storage medium is required to ensure inference efficiency due to the frequent data swapping between the storage medium and GPU.

\section*{Acknowledgments}
This work is partly supported by the Natural Science Foundation of China under Grant (No. 62206042, No. 62137001, and No. 62272093), the Joint Funds of Natural Science Foundation of Liaoning Province (No. 2023-MSBA-081), and the Fundamental Research Funds for the Central Universities under Grant (No. N2416012).

%% file: section/7_appednix.tex
\section{Appendix}
\subsection{License}
The terms of use for ClueWeb22 can be found on the Lemur Project website\footnote{\url{https://lemurproject.org/clueweb22}}, while CommonCrawl provides its terms of use on its official website\footnote{\url{https://commoncrawl.org/terms-of-use}}. All of these licenses and agreements allow their data for academic use.

\subsection{More Experimental Details of Overall Evaluation} \label{app:overall}
In this subsection, we describe further details of the implementation of overall evaluation. 

\textbf{Dataset.} We randomly selected about 8.28 million webpages from ClueWeb22-B English subset as the training set. To evaluate content extraction performance, we utilized a snapshot extracted from ClueWeb22-B, identified as \texttt{en0001-01}. This particular snapshot comprises 19,013 English webpages along with respective annotations. Notably, it's imperative to highlight that \texttt{en0001-01} was excluded from both the training, and validation datasets.

\textbf{Metrics.} In our experiments, we convert the web scanning task into a binary classification problem, so we can compute relevant metrics at the node level. However, some previous web scrapers would directly return the primary content without node information. Therefore, we directly check whether the reserved plain text contains the text spans of DOM tree nodes, which are annotated as ground truths in the benchmark.

\textbf{Computing Platform. } We conducted the training of \texttt{NeuScraper} on a server equipped with 8$\times$ NVIDIA A100-40G GPUs, with the training process spanning approximately 40 hours. For the evaluation of baseline scrapers, we utilized a setup comprising 2$\times$ Intel Xeon Gold-6348@2.60GHz CPUs with multiprocessing. In contrast, the evaluation of \texttt{NeuScraper} was carried out using 8$\times$ NVIDIA A100-40 GB GPUs, employing an inference batch size of 256 per GPU.

\subsection{More Experimental Details on Using Cleaned Web Data for Language Model Pretraining} 
\label{app:pretrain}
In this subsection, we describe additional details of the evaluation of the effectiveness of the cleaned web data in language model pretraining.

\textbf{Pretraining Corpus.} We utilize ClueWeb22-B and CommonCrawl \texttt{CC-MAIN-2023-50} as the source corpus for our pretraining endeavors. For ClueWeb22 , we employ various scrapers to acquire the corpus while ensuring an equivalent number of tokens, thereby pretraining the language model to mirror the performance of each scraper. For CommonCrawl, we used the pipeline from Pile-CC~\cite{gao2020pile}, but removed the language model filtering.  For various sizes of Pythia models, the corpus from ClueWeb22 consistently contains 13 billion tokens, while the corpus from Common Crawl is fixed at 2.8 billion tokens.

\textbf{Pretraining Details.}
Our pretraining framework extends from the \texttt{Lit-GPT}\footnote{\url{https://github.com/Lightning-AI/lit-gpt}} and we evaluate the performance of pretrained models using the standard \texttt{lm-evaluation-harness }toolkit\footnote{\url{https://github.com/EleutherAI/lm-evaluation-harness}}. Specifically, for all Pythia models, we employed the AdamW optimizer with a peak learning rate in line with \citet{biderman2023pythia}. The total batch size was set to 480 (with the batch size of 12 per GPU and gradient accumulation being set to 10). For ClueWeb22, the model undergoes training for just one epoch. For CommonCrawl, it is trained across three epochs due to the size of the corpus. All of the models were trained on 4$\times$ NVIDIA A100-40G GPUs. 

\textbf{Datasets for Evaluation. }
We choose 8 standard datasets to evaluate the performance of pretrained language models. Some of them are from the Pythia standard benchmark~\cite{biderman2023pythia}, supplemented by SWAG~\cite{zellers2018swag} and BLIMP~\cite{warstadt2020blimp}.

\textbf{Baselines. } In this experiments, we chose to use \texttt{htmlparser}\footnote{\url{https://htmlparser.sourceforge.net}} and \texttt{trafilatura}~\cite{barbaresi2021trafilatura} as the main baselines for comparison. \texttt{htmlparser} serves as the text pre-extraction tool for CommonCrawl WET file, while \texttt{trafilatura} has become the state-of-the-art web scraper.

\subsection{Performance on Multilingual Webpages}
\input{table/ml_scraping}
Thanks to the careful planning of ClueWeb22, which allows us to evaluate the performance of scrapers in different languages. Specifically, we tested on snapshots coded \texttt{0001-01} for each language, the results are shown in Table~\ref{tab:ml_scraping}. Among all the baseline scrapers, \texttt{NeuScraper} demonstrated excellent performance, even though it was trained only on English data.

\subsection{Case Study}
In this subsection, we show additional case studies of \texttt{NeuScraper} and \texttt{trafilatura}, our neural web scraper and a previously state-of-the-art web scraper. 

We first analyze the case in Figure~\ref{fig:case1}, where we use red boxes to indicate the content extracted by the scrapers. This is a college course page that contains some expertise in electrical engineering. When scraping this page, \texttt{trafilatura} loses a lot of textual content compared to our \texttt{NeuScraper}. By checking the raw HTML code, we found that there is an error caused by insufficient standardization of web pages: the paragraph tag ``\textit{<p>}'' is used for headings on this page instead of the standard ``\textit{<h>}'' tag. This page is readable for humans, but the HTML
 tag conveys an error that seriously affects the extraction performance of \texttt{trafilatura}. In contrast, our \texttt{NeuScraper} shows great adaptability. It not only extracts most of the paragraph content, but also removes useless information such as phone numbers, e-mails, dates, and so on.

Another typical case is interleaved boilerplate and body text, as shown in Figure~\ref{fig:case2}. We use blue boxes to indicate the content extracted by the scraper. In this case, the boilerplate and body text are written in the same way. The boilerplate also uses ``\textit{<h>}'' to identify headings and ``\textit{<p>}'' for paragraphs, instead of the list surrounded by ``\textit{<li>}'' in most cases. Recognizing it is difficult for \texttt{trafilatura}. \texttt{NeuScraper} leverages its ability to recognize latent semantic information to remove the boilerplate in such pages successfully. 
\clearpage
\newpage
\input{figure/case1}
\clearpage
\newpage
\input{figure/case2}

%% file: table/ml_scraping.tex
\begin{table*}[h!]
    \begin{center}
    \small
        \begin{tabular}[b]{lcccccccccc}
        \toprule
         & \multicolumn{2}{c}{\textbf{English}} & \multicolumn{2}{c}{\textbf{German}} & \multicolumn{2}{c}{\textbf{Spanish}} & \multicolumn{2}{c}{\textbf{French}} & \multicolumn{2}{c}{\textbf{Italian}}  \\
        \cmidrule(lr){2-3} \cmidrule(lr){4-5} \cmidrule(lr){6-7} \cmidrule(lr){8-9} \cmidrule(lr){10-11} 
               & \textbf{Acc} & \textbf{F1}  & \textbf{Acc} & \textbf{F1} & \textbf{Acc} & \textbf{F1} & \textbf{Acc} & \textbf{F1} & \textbf{Acc} & \textbf{F1}  \\
        \midrule
        bs4&41.29&58.10&40.49&57.23&39.34&56.18&40.05&56.84&38.92&55.72\\
        html2text&40.44&53.88&38.91&52.51&37.19&50.28&38.65&51.72&37.57&50.20\\
        boilerpipe&66.48&46.16&66.38&43.63&70.04&51.74&67.83&46.56&69.85&50.56\\
        jusText&62.58&22.17&65.84&42.98&61.25&2.13&60.79&2.63&61.56&0.53\\
        lxml&64.62&44.78&63.47&43.07&67.45&48.82&65.32&45.44&67.12&48.61\\
        inscriptis&45.35&58.98&43.82&57.27&42.74&56.30&42.99&56.19&43.42&56.49\\
        readability&68.47&48.22&70.16&50.17&72.08&54.38&71.10&52.21&72.69&54.85\\
        trafilatura&70.70&61.08&73.84&62.43&73.93&62.14&73.60&62.20&74.49&62.87\\ \midrule
        NeuScraper&86.35&83.90&79.10&73.02&78.89&71.90&76.58&68.12&78.76&71.33\\
        \midrule
        & \multicolumn{2}{c}{\textbf{Chinese}} & \multicolumn{2}{c}{\textbf{Japanese}} & \multicolumn{2}{c}{\textbf{Dutch}} & \multicolumn{2}{c}{\textbf{Portuguese}} & \multicolumn{2}{c}{\textbf{Polish}}   \\
        \cmidrule(lr){2-3} \cmidrule(lr){4-5} \cmidrule(lr){6-7} \cmidrule(lr){8-9} \cmidrule(lr){10-11} 
               & \textbf{Acc} & \textbf{F1}  & \textbf{Acc} & \textbf{F1} & \textbf{Acc} & \textbf{F1} & \textbf{Acc} & \textbf{F1} & \textbf{Acc} & \textbf{F1}  \\
        \midrule
        bs4&49.10&65.33&49.95&65.75&36.86&53.51&40.39&57.24&36.95&53.60\\
        html2text&48.29&63.94&50.00&64.74&35.44&48.96&38.57&52.09&36.16&49.26\\
        boilerpipe&61.31&42.44&57.33&30.26&70.01&44.82&67.93&49.14&66.96&36.91\\
        jusText&51.38&0.75&51.26&0.49&65.11&12.84&60.33&3.03&63.60&0.76\\
        lxml&62.22&52.79&60.38&50.16&66.16&41.36&66.59&48.73&65.72&40.01\\
        inscriptis&53.09&66.35&53.76&66.57&40.11&53.69&44.01&57.65&40.51&53.15\\
        readability&67.45&56.61&64.64&50.14&71.54&47.03&70.60&53.26&66.81&42.38\\
        trafilatura&68.57&63.29&71.82&67.08&74.06&59.88&72.64&61.67&71.58&53.02\\ \midrule
        NeuScraper&74.76&73.99&74.01&73.80&77.70&68.13&77.48&71.28&75.84&64.61\\
        
        \bottomrule
        \end{tabular}
        \caption{Scarping Performance in  Different Languages. We tested it on ClueWeb22 in different languages and NeuScraper showed significant improvements over the baseline scrapers. \label{tab:ml_scraping}}
    \end{center}
\end{table*}

%% file: figure/case1.tex
\begin{figure*}
\centering
\subfigure[Trafilatura.]{
\begin{minipage}[b]{1\textwidth}
\centering
\includegraphics[width=1\textwidth]{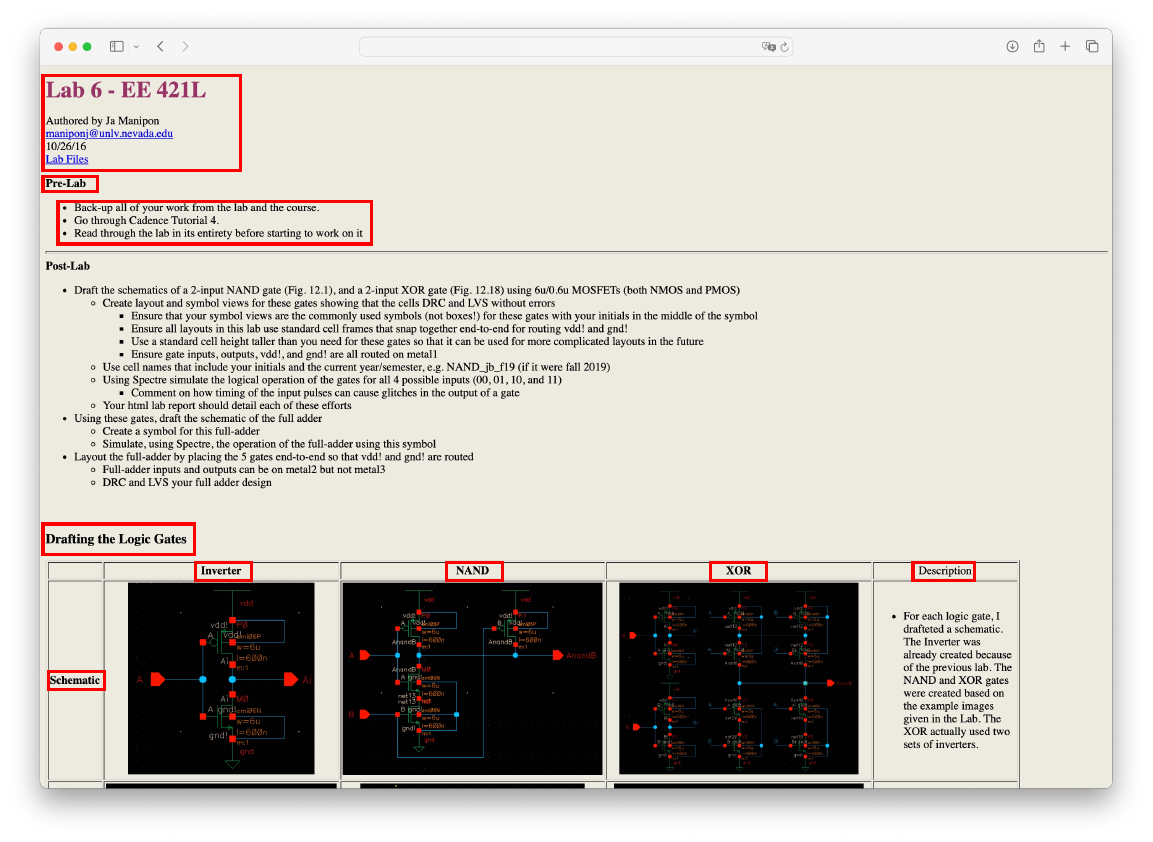}
\end{minipage}
}
\subfigure[NeuScraper.]{
\begin{minipage}[b]{1\textwidth}
\centering
\includegraphics[width=1\textwidth]{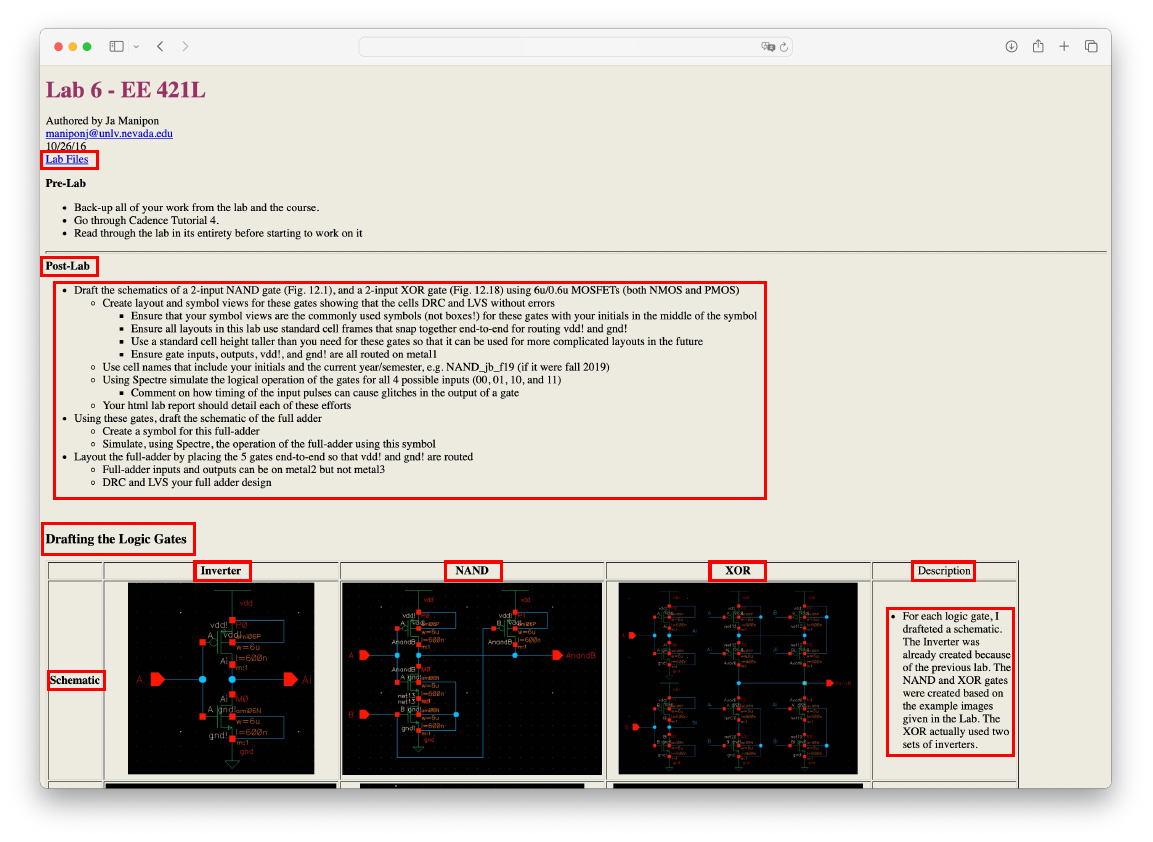}
\end{minipage}
}
\caption{Case\#1 of the Primary Content Extraction Results Using Different Scrapers. The extracted parts are highlighted with red boxes.} \label{fig:case1}
\end{figure*}

%% file: figure/case2.tex
\begin{figure*}
\centering
\subfigure[Trafilatura.]{
\begin{minipage}[b]{1\textwidth}
\centering
\includegraphics[width=1\textwidth]{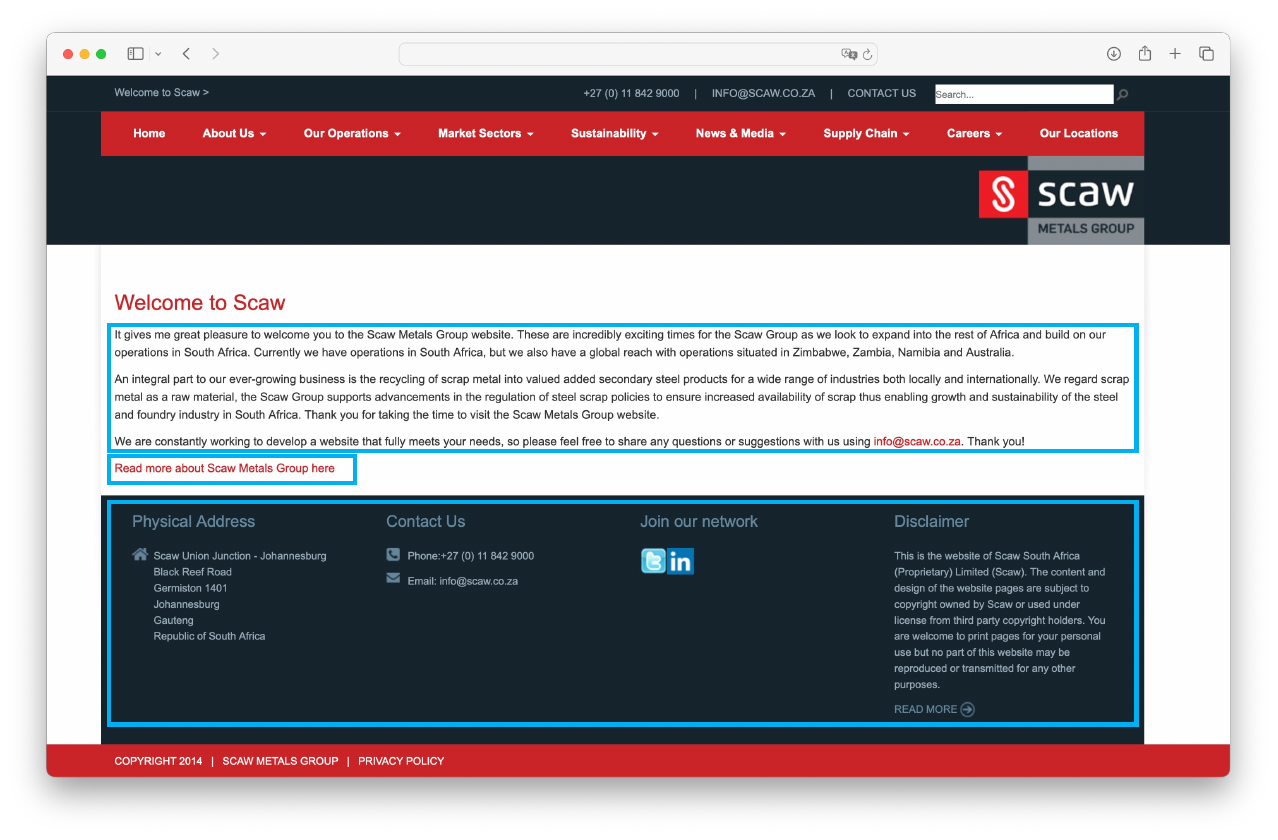}
\end{minipage}
}
\subfigure[NeuScraper.]{
\begin{minipage}[b]{1\textwidth}
\centering
\includegraphics[width=1\textwidth]{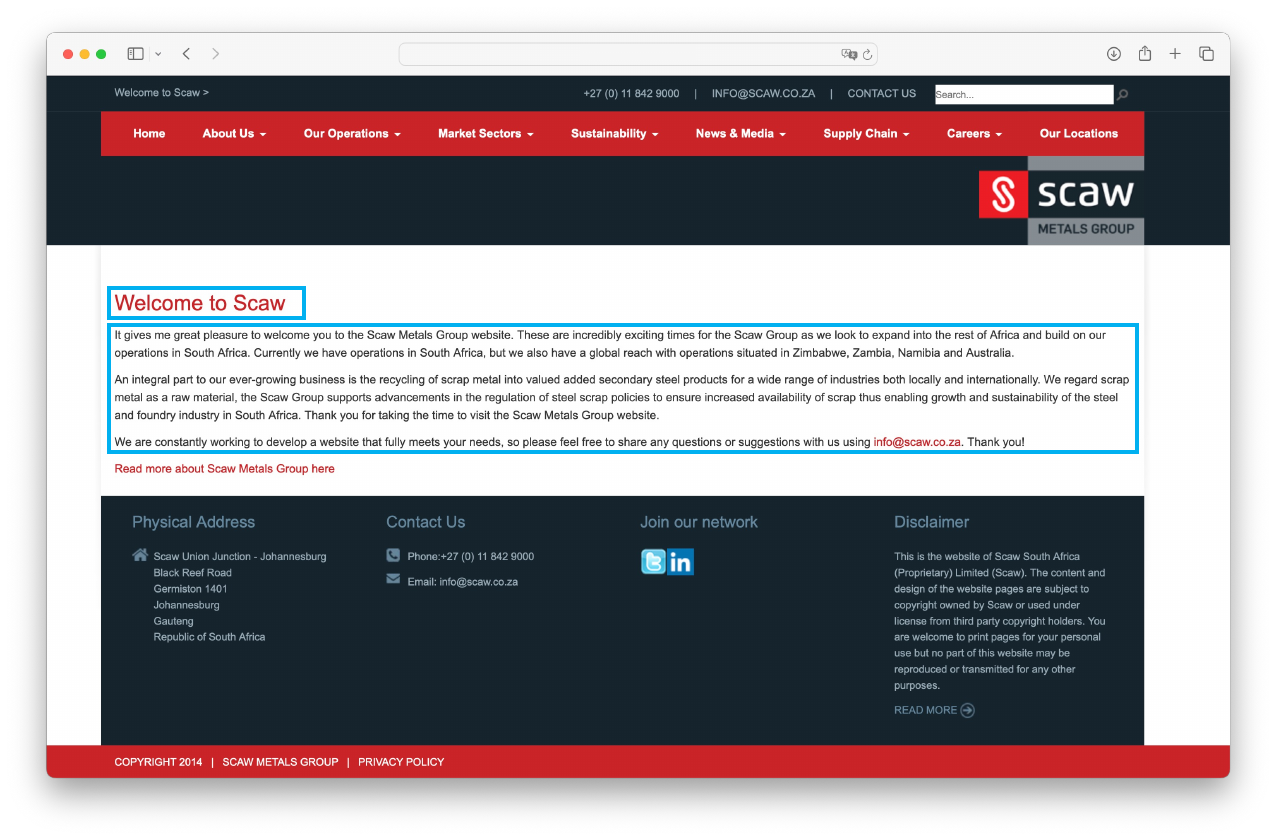}
\end{minipage}
}
\caption{Case\#2 of the Primary Content Extraction Results Using Different Scrapers. The extracted parts are highlighted with blue boxes.} \label{fig:case2}
\end{figure*}